%% file: arxiv.tex
\documentclass[11pt]{article}

\usepackage[preprint]{acl}
\usepackage{times}
\usepackage{latexsym}
\usepackage[T1]{fontenc}
\usepackage[utf8]{inputenc}
\usepackage{microtype}
\usepackage{inconsolata}
\usepackage{graphicx}
\usepackage{booktabs}
\usepackage{amsmath}
\usepackage{amssymb}
\usepackage{xcolor}
\usepackage{titletoc}

\newcommand{\method}{\textsc{Jev-DLC}}
\newcommand{\benchmark}{\textsc{TCPA-Bench}}
\newcommand{\jone}{\textsc{JEV-SFT}}
\newcommand{\appref}[1]{\ref{#1}}

\title{Jev-DLC: Decision-Level Continual Learning from Typed Feedback}

\author{
  Yangyang Feng$^{1,\dagger}$ \quad
  Jianle Gan$^{1,\dagger}$ \quad
  Zhuoyan Feng$^{2}$ \quad
  Chengwei Qin$^{1,*}$ \\
  $^{1}$The Hong Kong University of Science and Technology (Guangzhou) \\
  $^{2}$Sun Yat-sen University \\
  \texttt{yfeng044@connect.hkust-gz.edu.cn} \quad
  \texttt{jgan307@connect.hkust-gz.edu.cn} \\
  \texttt{fengzhy28@mail2.sysu.edu.cn} \quad
  \texttt{chengweiqin@hkust-gz.edu.cn} \\
  $^{\dagger}$Equal contribution \quad $^{*}$Corresponding author
}

\begin{document}
\maketitle

\begin{abstract}
\input{sections/00-abstract}
\end{abstract}

\input{sections/01-introduction}
\input{sections/02-related-work}
\input{sections/03-problem-setup}
\input{sections/04-method}
\input{sections/05-interface-analysis}
\input{sections/06-benchmark}
\input{sections/07-experiments}
\input{sections/08-conclusion}
\clearpage
\input{sections/09-limitations}

\bibliography{references}

\clearpage
\appendix
\raggedbottom
\input{sections/appendix/00-contents}
\clearpage
\input{sections/appendix/a-notation-and-interface}
\input{sections/appendix/c-method-details}
\input{sections/appendix/b-theory-and-proofs}
\input{sections/appendix/d-benchmark-and-settings}
\input{sections/appendix/e-additional-results}
\input{sections/appendix/f-history-and-robustness}
\input{sections/appendix/g-prompts-api-and-cases}
\input{sections/appendix/h-reproducibility-and-scope}

\end{document}

%% file: sections/00-abstract.tex
Individual preferences often become visible only after deployment, when a user
accepts one decision or revokes it in favor of another. Jev-style typed
interfaces make this feedback explicit by identifying the selected action and,
under revocation, the decision it replaces. We study how these consequences
persist across requests as Decision-Level Continual Learning from Typed
Feedback (Jev-DLC). Jev-DLC treats each correction as both evidence about a
recurring action identity and a directed relation between the preferred and
rejected choices. It preserves these signals in complementary bounded channels
above a frozen scorer and combines their scores to shape the next decision.
The method retains no raw interaction history, adds no scorer call, and performs
no runtime language-model backward pass. We also introduce \benchmark{}, a
chronological suite that credits a correction only through its effect on later
decisions. On TCPA-Repeat, Jev-DLC reaches 44.83\% user-macro accuracy overall,
2.43 points above Recent-4 history replay. After at least two earlier feedback
events, it reaches 56.07\%, a 7.54-point margin over the strongest reported comparison
at this depth. These results show how explicit typed corrections
can become compact causal state for continual personalization.

%% file: sections/01-introduction.tex
\section{Introduction}
\label{sec:introduction}

\begin{figure}[t]
  \centering
  \includegraphics[width=0.98\linewidth]{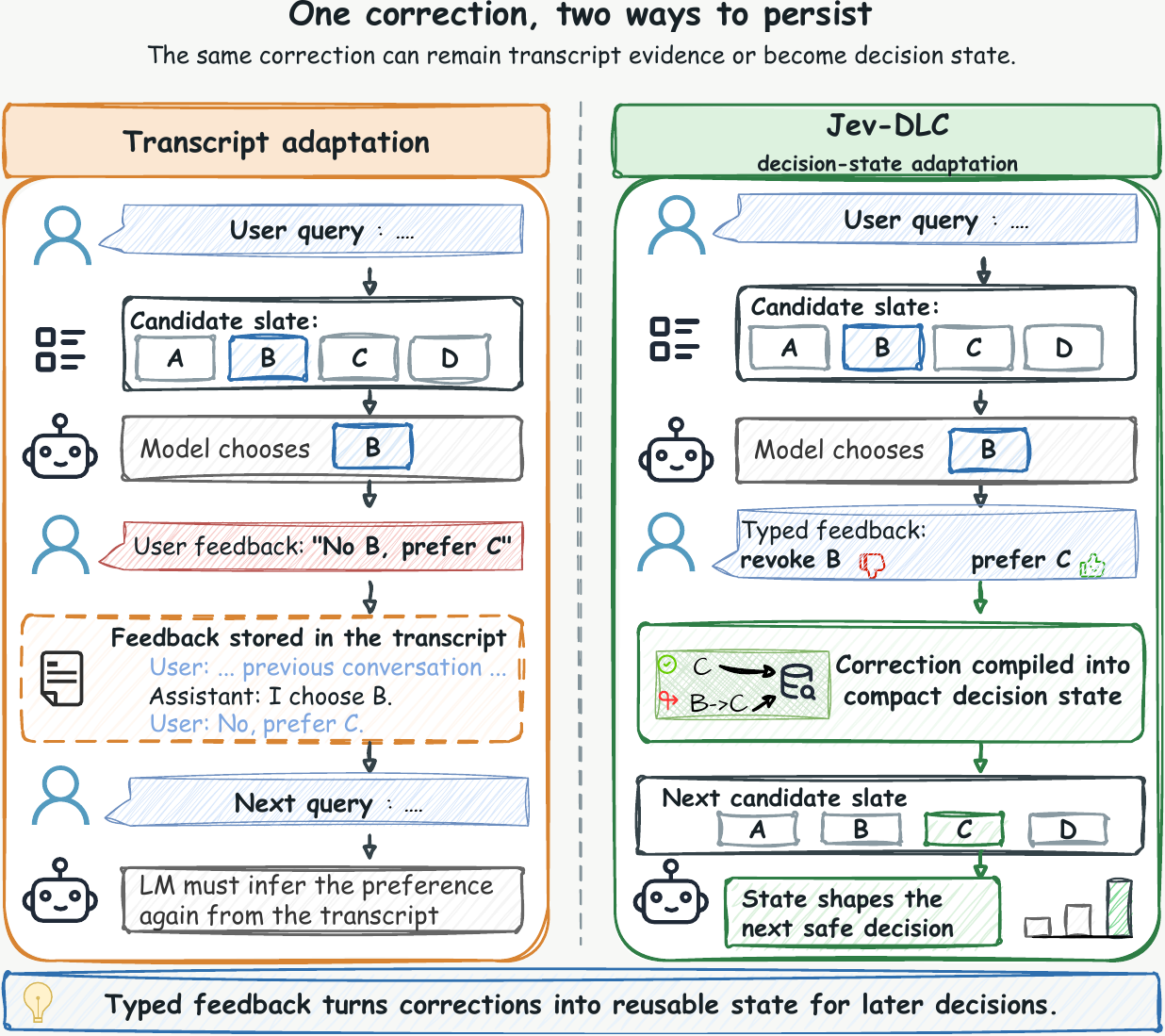}
  \caption{A typed correction exposes two forms of persistence. The selected
  action contributes evidence about what may recur, while the preferred--rejected
  pair gives a direction for repairing later decisions. Jev-DLC combines both
  through complementary bounded channels above a frozen scorer.}
  \label{fig:intro}
\end{figure}

Individual preference often becomes visible only after deployment. A user may
accept one choice, replace another, or correct a system only after seeing its
decision. Personalization methods use this information by retrieving or
replaying past interactions, or by adapting a user-specific representation or
model from accumulated behavior
\citep{lau2024personalized,li2024personalized,salemi2024lamp}. These approaches
preserve rich evidence, while later predictions must reinterpret that evidence
or consult additional mutable parameters. A typed interface offers another
possibility. When feedback already identifies the accepted action and the one
it replaces, the system can preserve the consequence of the correction rather
than reconstruct it from the original interaction.

Jev-style systems make the consequence of a correction explicit. Given a finite
set of candidate actions, the model commits a typed decision that the user can
accept or revoke in favor of a named alternative
\citep{typesafe2026api,almeida2026jev}. Jev supplies the judgment for the current
request, but the interface leaves open how that correction should affect the
next one. Jev-DLC addresses this persistence problem above a frozen typed
scorer. A separate per-user state carries the consequences of earlier
feedback into the next decision.

A typed correction exposes two forms of persistence. The identity of the
preferred action records what the user chose and can reveal a categorical
tendency when actions recur. The preferred--rejected pair records how one
decision should replace another and defines a local direction that can transfer
to related candidate slates. Jev-DLC preserves both through a conservative
channel that combines repeated action evidence with local repair and a
responsive channel that carries correction direction without the count prior.
Their combined scores allow earlier feedback to influence the next decision
while the frozen scorer continues to interpret its context and candidates.

Causal per-user weights combine the channels before each decision. After the
runtime commits its prediction, accept or revoke feedback updates both bounded
channels and their weights for the same user's next request. Each directional
update makes the smallest capped change needed to repair the exposed ordering,
while revealed channel losses adjust their relative contribution. The system retains no raw interaction history, adds no scorer
call, and performs no runtime language-model backward pass.

We evaluate this formulation with \benchmark{}, which counts a correction only
through its effect on later decisions. On TCPA-Repeat, the chronological
repeated-action track derived from LaMP-2 \citep{salemi2024lamp}, Jev-DLC reaches
44.83\% user-macro accuracy on the full stream, compared with 39.44\% for the
frozen scorer and 42.40\% for Recent-4 history replay. After at least two earlier
feedback events, it reaches 56.07\%, a 7.54-point margin over the strongest reported
comparison at this depth. The results connect the retained correction
state to later decision quality under a bounded serving budget. The experiments
also examine feedback depth, changing candidate items, and alternative fusion
strategies to characterize where this persistence is useful.

The paper makes three contributions.
\begin{itemize}
  \item We formulate typed continual preference alignment as a chronological
  decision problem in which accept and revoke feedback can affect only later
  decisions for the same user, and instantiate this contract in \benchmark{}.
  \item We develop Jev-DLC, which preserves recurring action evidence and
  correction direction in complementary bounded channels above a frozen scorer.
  Our analysis connects typed action sufficiency, local minimum-change repair,
  and the geometry through which a correction influences later decisions.
  \item We show how this compact state improves repeated-action decisions
  without raw-history replay or runtime language-model training. Comparisons
  with history replay, user encoding, temporal memory, and meta-adaptation
  characterize its accuracy and serving tradeoff under the same chronological
  feedback protocol.
\end{itemize}

%% file: sections/02-related-work.tex
\section{Related Work}
\label{sec:related-work}

\subsection{Personalization from User History}
Personalization systems often treat past behavior as evidence that must be
revisited. In-context and retrieval-based methods place previous interactions
back into the prompt, while other approaches learn user representations or
personalized parameters \citep{jang2023personalized,rame2023rewarded,
siththaranjan2023distributional,lau2024personalized,li2024personalized}.
LaMP makes this history-based formulation explicit across classification and
generation tasks \citep{salemi2024lamp}; MIND provides a large sequential
recommendation setting built from news-click histories \citep{wu2020mind}.
Recent-K is the closest system comparison for our setting because it preserves
recent evidence and asks the same frozen scorer to reinterpret it at every
event, following the history-based personalization protocol used by LaMP
\citep{salemi2024lamp}. Jev-DLC studies the setting in which feedback has
already exposed the decision relation and therefore lets that relation persist
as a direct influence on future decisions rather than retaining the full
evidence that produced it. Existing work thus solves how to infer a user's
preference from past behavior; we study how an explicit typed correction
persists after that inference has already occurred.

\subsection{Continual and Online Adaptation}
Continual-learning methods manage the stability--plasticity trade-off through
replay, regularization, parameter isolation, or dynamic capacity
\citep{kirkpatrick2017overcoming,lopezpaz2017gradient,parisi2019continual,
delange2022continual}. Language-model studies apply related ideas to sequential
tasks, instructions, and knowledge updates \citep{jang2022continual,
scialom2022finetuned}. These methods make a model, adapter, or shared memory
mutable as new information arrives. Jev-DLC instead freezes the scorer and
restricts mutability to a per-user decision state above it. Replay, subspace,
first-order, and second-order learners remain useful comparisons because they
represent different choices about what is allowed to change and what must be
retained. Jev-DLC places the stability--plasticity trade-off at the level of a
per-user decision state and uses PA-I as its minimum-change update.
Existing work therefore studies how a mutable model remains useful across
changing tasks; we study how a frozen scorer can adapt for one user through a
small persistent decision state.

\subsection{Typed Decisions and Online Decision Updates}
Jev-style systems expose finite candidate choices and read a typed decision from
the model output \citep{openjev2026modelcard}. This interface makes an accept or
revoke an explicit decision-level correction, but it does not by itself define
how that correction should alter later requests. Pairwise preference models
provide the corresponding relative-decision language
\citep{bradley1952rank}. Passive-aggressive learning supplies a minimum-change
online update with an optional slack cap \citep{crammer2006passive}, while
adaptive-gradient and online Newton methods provide alternative update
geometries \citep{duchi2011adaptive,shalevshwartz2012online}. Jev-DLC uses PA-I
to connect an explicit accept/revoke interface to that update under a frozen
typed scorer and a chronological per-user serving contract, then tests whether
the resulting state changes later decisions. Conversation benchmarks and
preference arenas evaluate broader response quality without making this state
transition the object of evaluation
\citep{zheng2023judging,chiang2024chatbot}. Existing work separately provides a
typed readout or a minimum-change online update; we connect the two layers and
measure the resulting adaptation trajectory.

%% file: sections/03-problem-setup.tex
\section{Problem Setup}
\label{sec:problem-setup}

Continual preference alignment studies whether a user's typed correction can
improve later decisions without changing the scorer or replaying the
interaction. We make the information boundary explicit before describing any
particular update rule or evaluation set.

\subsection{Ordered Typed Interaction}
For user $u$, events arrive in chronological order as
$E_{u,1},\ldots,E_{u,T_u}$. Event $E_{u,t}$ contains a context $x_t$ and a
finite candidate slate $\mathcal A_t$. A frozen scorer assigns a score to each
candidate, and a fixed mask $M_t(a)$ removes actions that violate hard
constraints. The learner predicts from the safe slate
$\mathcal A_t^{\mathrm{safe}}$ and commits a decision $\hat a_t$ before any
feedback is available. The committed prediction is the object evaluated at
event $t$.

\subsection{Typed Feedback}
After the prediction has been committed, the interface emits either
$\operatorname{accept}(\hat a_t)$ or
$\operatorname{revoke}(\hat a_t\!\rightarrow a_t^\star)$. Acceptance confirms
the committed action. Revocation identifies both the rejected action
$a_t^- = \hat a_t$ and its replacement $a_t^+=a_t^\star$. Thus feedback
supervises a decision that has already been made and, in the revoke case,
specifies the replacement relation directly. It is decision-level supervision,
not an unconstrained free-text critique.

\subsection{Persistent State and Causality}
The learner maintains a state $\mu_{u,t}$ for each user and updates it with the
current event and the feedback revealed after prediction,
\begin{equation}
 \mu_{u,t+1}=U(\mu_{u,t},E_{u,t},F_t).
 \label{eq:update-contract}
\end{equation}
The prediction at event $t$ may depend on feedback from events $1{:}t-1$ only.
The feedback $F_t$ can affect events $t+1{:}T_u$, but cannot change the
prediction that exposed it. State is reset between users and is subject to a
bounded-state serving constraint. For a fixed user, we write this state as
$\mu_t$. The main state-based path retains neither raw previous interactions
nor candidate text, and scorer parameters remain fixed at runtime. The scorer
evaluates the current slate; $\mu_{u,t}$ carries only the user's accumulated
consequences of earlier typed feedback.

\subsection{Evaluation Objective}
The central question is whether past typed feedback improves later decisions
for the same user. The objective is therefore to evaluate the state induced by
the prefix of a stream, rather than to refit the event whose target has just
been revealed. User-macro accuracy is the primary metric because the unit of
personalization is the user. Other statistical inference details and serving
measurements are reported with the experiments and in
Appendix~\appref{app:metrics}.

%% file: sections/04-method.tex
\section{Method}
\label{sec:method}

Jev-DLC separates current-request interpretation from persistent preference.
A frozen typed scorer supplies the base judgment for the current candidate
slate. Two bounded user-state channels preserve the recurring action evidence
and correction direction exposed by earlier feedback. Their combined scores
let these consequences shape the next decision without revisiting the original
interactions.

\subsection{Frozen Typed Decision Layer}

Given a rendered context and a candidate slate, \jone{} scores the option tokens
at the first output position. Each safe candidate $a$ receives a base logit
$b_t(a)$ and a frozen hidden representation $h_t(a)\in\mathbb R^D$. A shared
linear map $P\in\mathbb R^{d\times D}$ projects the representation into the
state space,
\begin{equation}
 r_t(a)=\frac{Ph_t(a)}{\lVert Ph_t(a)\rVert_2},\qquad d=32.
 \label{eq:representation}
\end{equation}
The map is fitted once on chronological streams from training users, shared by
all users, and frozen before development and test replay. The scorer parameters,
projection, candidate order, and hard mask remain fixed at runtime. Alternative
representation maps are reported in
Appendix~\appref{app:representation-analysis}.

\subsection{A Typed Correction Provides a Prior and a Direction}

The runtime commits $\hat a_t$ before feedback is revealed. A revoke identifies
the rejected prediction and its replacement. An accept confirms the committed
action, which is paired with its highest-probability safe competitor. We denote
the preferred and paired actions by $a_t^+$ and $a_t^-$, respectively. The
correction then provides two related signals.

First, the preferred action contributes evidence to an observable semantic
identity $c(a_t^+)$. Let $n_t(k)$ be the number of previous corrections assigned
to semantic identity $k$, let $N_t=\sum_k n_t(k)$, and let $K_t$ be the number
of semantic identities represented in the current slate. The resulting
smoothed prior adjustment is
\begin{equation}
 q_t(a)=\lambda_c\log
 \frac{n_t(c(a))+\alpha}{N_t+\alpha K_t}.
 \label{eq:semantic-prior}
\end{equation}
Second, the same correction identifies a directed relation in the frozen
candidate space,
\begin{equation}
 d_t=r_t(a_t^+)-r_t(a_t^-),\qquad
 \beta_t=b_t(a_t^+)-b_t(a_t^-).
 \label{eq:typed-contrast}
\end{equation}
The prior records what the user repeatedly chooses. The contrast records how
one committed decision should be replaced by another.

\subsection{Conservative and Responsive Channels}

The conservative channel combines the semantic prior with a local correction
vector $\mu_t^{\mathrm c}\in\mathbb R^d$. Its score is
\begin{equation}
 \begin{aligned}
 x_t^{\mathrm c}(a)&=\lambda_p r_t(a),\\
 s_t^{\mathrm c}(a)&=b_t(a)+q_t(a)
 +{\mu_t^{\mathrm c}}^\top x_t^{\mathrm c}(a).
 \end{aligned}
 \label{eq:conservative-score}
\end{equation}
The responsive channel omits the count prior and stores the correction geometry
directly in $\mu_t^{\mathrm r}\in\mathbb R^d$,
\begin{equation}
 \begin{aligned}
 x_t^{\mathrm r}(a)&=r_t(a),\\
 s_t^{\mathrm r}(a)&=b_t(a)+{\mu_t^{\mathrm r}}^\top x_t^{\mathrm r}(a).
 \end{aligned}
 \label{eq:responsive-score}
\end{equation}
Each channel yields a distribution
$p_t^j=\operatorname{softmax}_{a\in\mathcal A_t^{\mathrm{safe}}}s_t^j(a)$ for
$j\in\{\mathrm c,\mathrm r\}$.

After feedback, both channels receive the pair elicited by the same committed
mixture decision. Their contrasts and base margins are
\begin{equation}
 \begin{aligned}
 d_t^j&=x_t^j(a_t^+)-x_t^j(a_t^-),\\
 \beta_t^{\mathrm c}&=b_t(a_t^+)+q_t(a_t^+)
 -b_t(a_t^-)-q_t(a_t^-),\\
 \beta_t^{\mathrm r}&=b_t(a_t^+)-b_t(a_t^-).
 \end{aligned}
 \label{eq:channel-contrasts}
\end{equation}
For $j\in\{\mathrm c,\mathrm r\}$, the margin violation and minimum-change
transition are
\begin{equation}
 \begin{aligned}
 \ell_t^j&=\left[\gamma-\beta_t^j
 -{\mu_t^j}^\top d_t^j\right]_+,\\
 \mu_{t+1}^j&=\mu_t^j+\tau_t^j d_t^j.
 \end{aligned}
 \label{eq:channel-hinge}
\end{equation}
\begin{equation}
 \tau_t^j=\min\!\left(C,\frac{\ell_t^j}{\lVert d_t^j\rVert_2^2}\right),
 \label{eq:jev-dlc-update}
\end{equation}
with $\tau_t^j=0$ when $d_t^j=0$. If a channel already respects the correction,
it remains unchanged. Otherwise, it moves only as far as needed to repair the
ordering, subject to the cap. The semantic count is incremented only after the
current decision has been evaluated. We use $\gamma=C=1$ and freeze all channel
parameters before test replay.

\subsection{Combining the Two Channels}

The two channels need not be equally reliable for every user or at every point
in a stream. Jev-DLC therefore maintains two log weights
$g_t^{\mathrm c}$ and $g_t^{\mathrm r}$. Before event $t$, their normalized
values define the responsive weight
\begin{equation}
 \rho_t=
 \frac{\exp(g_t^{\mathrm r})}
 {\exp(g_t^{\mathrm c})+\exp(g_t^{\mathrm r})}.
 \label{eq:reliability-weight}
\end{equation}
The runtime combines the expert distributions geometrically,
\begin{equation}
 \begin{aligned}
 \bar p_t(a)&=p_t^{\mathrm c}(a)^{1-\rho_t}
 p_t^{\mathrm r}(a)^{\rho_t},\\
 p_t(a)&=\frac{\bar p_t(a)}
 {\sum_{a'}\bar p_t(a')},\\
 \hat a_t&=\arg\max_a p_t(a).
 \end{aligned}
 \label{eq:reliability-mixture}
\end{equation}
Only after the target $a_t^+$ is revealed does the runtime observe each
channel's zero-one loss
$L_t^j=\mathbf{1}[\arg\max_a p_t^j(a)\ne a_t^+]$. It applies the exponential
update $\tilde g_{t+1}^j=g_t^j-\eta L_t^j$, normalizes the two weights, and
mixes a fixed share $\sigma$ back toward the uniform distribution,
\begin{equation}
 \begin{aligned}
 \bar w_{t+1}^j&=\frac{\exp(\tilde g_{t+1}^j)}
 {\sum_k\exp(\tilde g_{t+1}^k)},\\
 w_{t+1}^j&=(1-\sigma)\bar w_{t+1}^j+\frac{\sigma}{2},\\
 g_{t+1}^j&=\log w_{t+1}^j.
 \end{aligned}
 \label{eq:fixed-share-update}
\end{equation}
This update lets reliability change as evidence accumulates while preventing
either channel from becoming permanently unreachable. The initialization and
frozen configuration are given in Appendix~\appref{app:method-details}.

\subsection{Serving Contract and State Budget}

At each event, one frozen scorer call produces the candidate logits and
representations. The two channels form their distributions from the current
user state, the reliability weights mix them, and the runtime commits the
result. The revealed accept or revoke then updates the reliability weights,
the two correction vectors, and the semantic count state for the user's next
event. Every update uses the feedback produced by the committed mixture rather
than a counterfactual decision from either channel.

The complete state is
$z_{u,t}=(n_{u,t},\mu_{u,t}^{\mathrm c},\mu_{u,t}^{\mathrm r},
g_{u,t}^{\mathrm c},g_{u,t}^{\mathrm r})$. Its maximum observed size is
324 bytes on TCPA-Repeat and 320 bytes on TCPA-Open. Jev-DLC retains no raw
interaction history, adds no scorer forward pass, and performs no runtime
language-model backward pass. Resetting the counts, correction vectors, and
reliability weights restores the frozen scorer over the same safe slate.
The runtime interface, state accounting, and frozen configuration are detailed
in Appendix~\appref{app:method-details}.

\begin{figure*}[t]
  \centering
  \includegraphics[width=\textwidth]{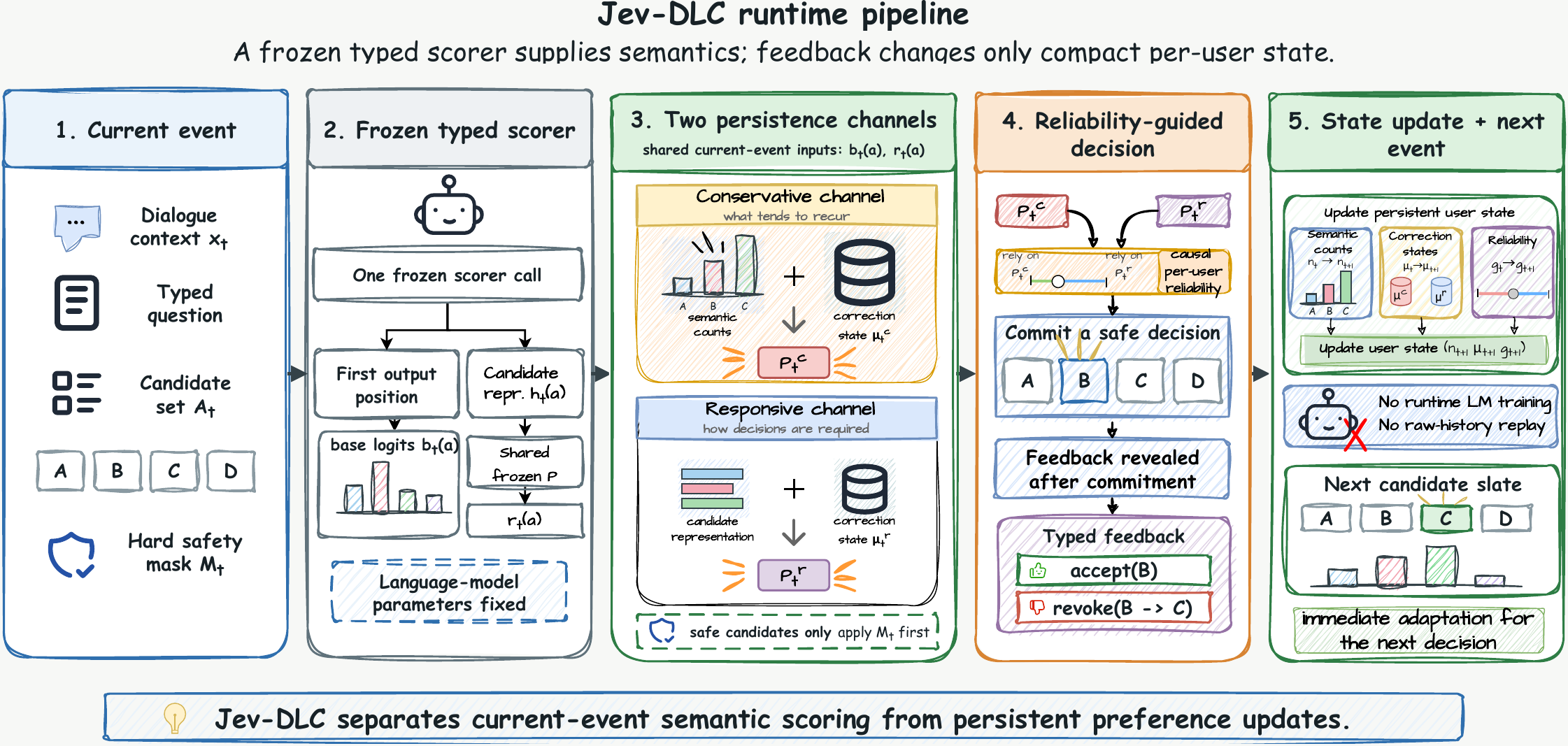}
  \caption{Typed feedback becomes persistent state above a frozen scorer.
  Conservative and responsive channels retain recurring action evidence and
  correction geometry. Causal per-user weights combine their distributions
  before commitment, and revealed feedback updates the channels and weights
  for later requests.}
  \label{fig:method-flow}
\end{figure*}

%% file: sections/05-interface-analysis.tex
\section{Theory}
\label{sec:analysis}

The analysis explains what a typed correction preserves and how that
information can influence a later decision. Action sufficiency identifies when
the selected action retains the preference information in its realization.
Minimum-change repair then converts the exposed ordering into a local state
transition, whose candidate-space geometry determines transfer to future pairs.

\subsection{What Typed Feedback Preserves}

A typed action can replace its language realization when the surface form adds
no preference information after the action is known. Let realization $R$ map
deterministically to action $A=g_x(R)$, and suppose
\begin{equation}
 p_{w,\zeta}(R\mid x)=q_\zeta(R\mid A,x)p_w(A\mid x),
 \label{eq:realization-factorization}
\end{equation}
where $q_\zeta$ is independent of preference parameter $w$.

\paragraph{Proposition 1 (action sufficiency).}
Under Eq.~\ref{eq:realization-factorization}, $A$ is sufficient for $w$
conditional on $x$. For any square-integrable realization-dependent update
$V(R,F)$ and $\bar V(A,F)=\mathbb E[V\mid A,F]$,
\begin{equation}
 \operatorname{Cov}(V)=\operatorname{Cov}(\bar V)
 +\mathbb E[\operatorname{Cov}(V\mid A,F)].
 \label{eq:rb-covariance}
\end{equation}

\paragraph{Proof sketch.}
The likelihood depends on $w$ only through $p_w(A\mid x)$, so the factorization
criterion gives sufficiency. The second statement is the law of total
covariance. Its final term is positive semidefinite and measures variation
among realizations of the same typed action. Typed conditioning therefore
removes realization nuisance while retaining the preference information
specified by the factorization. Appendix~\appref{app:theory} gives the complete
sequence-level result.

\subsection{Two Persistence Geometries}

The action identity and pairwise direction influence future decisions in
different ways. The semantic term in Eq.~\ref{eq:semantic-prior} is a
Dirichlet-smoothed log-frequency score. Up to a candidate-independent constant,
it increases with the number of earlier corrections selecting the same semantic
identity. It therefore carries repeated categorical tendency without retaining
the events that produced the counts.

The pairwise signal instead defines, for each channel $j$, a feasible half-space
\begin{equation}
 \mathcal H_t^j=\{\mu\colon
 \beta_t^j+\mu^\top d_t^j\ge\gamma\}.
 \label{eq:typed-halfspace}
\end{equation}

\paragraph{Proposition 2 (local minimum-change repair).}
For $d_t^j\ne0$, Eq.~\ref{eq:jev-dlc-update} is the unique PA-I state
transition for the exposed ordering. If the state already lies in
$\mathcal H_t^j$, it remains unchanged. Otherwise, an uncapped step is the
Euclidean projection onto the boundary; the cap limits the same directed move.

\paragraph{Proof sketch.}
KKT stationarity makes the minimizing displacement parallel to $d_t^j$.
Complementary slackness gives coefficient
$\ell_t^j/\lVert d_t^j\rVert_2^2$, clipped at $C$, and strict convexity gives
uniqueness. Thus the correction changes no state direction orthogonal to
$d_t^j$. The complete derivation and comparator-relative violation bound appear
in Appendix~\appref{app:theory}.

The geometry determines transfer to a later event $s>t$:
\begin{equation}
 {d_s^j}^\top(\mu_{t+1}^j-\mu_t^j)
 =\tau_t^j {d_s^j}^\top d_t^j.
 \label{eq:future-contrast-transfer}
\end{equation}
A later pair inherits the repair according to its alignment with the earlier
correction. If $\lVert x_t^j(a)\rVert_2\le\kappa_j$, then
\begin{equation}
 \lVert\mu_{t+1}^j-\mu_t^j\rVert_2\le2C\kappa_j,
 \qquad \kappa_{\mathrm c}=\lambda_p,\quad
 \kappa_{\mathrm r}=1.
 \label{eq:step-bound}
\end{equation}
The state remains fixed-dimensional, while the prior and direction encode
respectively what recurs and which candidate relation was corrected.

\subsection{From Local Repair to Persistent Decisions}

The mixture in Eq.~\ref{eq:reliability-mixture} brings the two persistence
geometries into one decision. The count term carries evidence across events
sharing an action identity, while the correction vectors transfer repair
according to candidate-space alignment. Because the conservative margin
includes the prior and the responsive margin does not, the same feedback can
produce different updates in the two channels. The user state preserves both
trajectories, and the pre-decision weights determine their contribution.

Together, action sufficiency and local repair explain how a correction can be
retained without its original language realization and reused on a later slate.
The experiments measure the resulting decision quality.
Appendix~\appref{app:theory} gives the channel-weighting analysis;
Appendix~\appref{app:fusion-strategies} reports other implementations of the
same two-channel fusion.

%% file: sections/06-benchmark.tex
\section{TCPA: Chronological Evaluation of Typed Adaptation}
\label{sec:benchmark}

TCPA evaluates whether typed feedback changes later decisions for the same user.
Every track follows one causal serving contract. The runner predicts and
commits, the evaluator then reveals typed feedback, the runner updates the
user state, and only subsequent decisions receive the benefit of that update.
This ordering separates continual adaptation from refitting the event that
revealed its target.

\subsection{Predict-Then-Update Protocol}

For each user, events remain in source order and users are disjoint across
training, development, and test. At event $t$, the runner sees the current
context and candidate slate, applies the frozen hard mask, and commits a
prediction before the evaluator exposes the target or feedback. The evaluator
returns either an acceptance or a revocation with its replacement. The runner
then updates the user's persistent state. The state is reset when a new user
stream begins, and event $t$ is scored before its feedback can affect event
$t+1$ or any later event. Targets are isolated from the serving process until
the committed prediction has been recorded.
Each scored event exposes the correction prefix available before commitment.
Any measured gain can therefore be attributed to information genuinely
available online.

\subsection{TCPA-Repeat}

TCPA-Repeat is our chronological repeated-action personalization track derived
from LaMP-2. It presents repeated decisions under a stable finite ontology and
preserves chronological user streams with enough repeated interaction to test
whether accumulated feedback improves later decisions. This setting makes the
central continual-adaptation question observable because the corrected action
can reappear after the correction. The converter, label mapping, filtering
rules, hashes, and split construction are specified in
Appendix~\appref{app:repeat-construction}.

\subsection{Additional Diagnostic Tracks}

TCPA-Open is a changing-item weak-transfer boundary track derived from MIND.
TCPA-Dynamics provides controlled stability, shift, recurrence, and noise
diagnostics. Together, these tracks complement TCPA-Repeat by probing transfer
across changing items and controlled temporal regimes. Their full construction
and results appear in Appendices~\appref{app:diagnostic-construction}
and~\appref{app:diagnostic-results}.

User-macro accuracy is the primary metric. System accounting records persistent
state, retained history, and extra model computation so that accuracy and
serving cost are evaluated together. Bootstrap inference and the complete
metric definitions are provided in Appendix~\appref{app:metrics}; representation
analyses appear in Appendix~\appref{app:representation-analysis}.

%% file: sections/07-experiments.tex
\section{Experiments}
\label{sec:experiments}

The experiments test whether persistent typed state improves later decisions
and what that improvement requires at serving time. TCPA-Repeat evaluates
reuse across recurring actions; feedback-depth strata show the decisions made
after different amounts of prior feedback. All methods use the frozen JEV-SFT
decision layer, whose current-event anchors appear in
Appendix~\appref{app:scorer-anchors}. The main tables report one Jev-DLC
implementation. Complete channel and fusion results appear in
Appendix~\appref{app:fusion-strategies}.

\subsection{Later-Decision Accuracy}

The comparisons span different ways of carrying user feedback into the next
decision. Recent-4 retains four typed interactions and asks JEV-SFT to
reinterpret them in an enlarged prompt. PPlug encodes user behavior, TAP-PER
queries bounded temporal memory, and MRM adapts a low-dimensional head. Our
TCPA instantiations of these three methods share Jev-DLC's frozen scorer cache
and learned candidate projection. All comparisons use the same safe slates,
targets, chronological order, and predict-then-update protocol. Shared
components are fitted on training users and configurations selected on
development users. Appendix~\appref{app:baseline-contracts} gives the source
method adaptations and representation provenance of the archived online
references, including AdaGrad.

\begin{table}[t]
\centering
\scriptsize
\setlength{\tabcolsep}{3.0pt}
\begin{tabular*}{\columnwidth}{@{\extracolsep{\fill}}lrrrr@{}}
\toprule
Method & Full & $d\ge2$ & $d\ge4$ & $d\ge8$ \\
\midrule
Frozen & 39.44 & 41.58 & 36.48 & 40.38 \\
Recent-4 & 42.40 & 46.40 & 37.56 & 42.17 \\
AdaGrad & 41.99 & 48.53 & 39.33 & 41.68 \\
PPlug & 40.37 & 41.77 & 37.86 & 43.38 \\
TAP-PER & 41.99 & 46.48 & 46.45 & 44.25 \\
MRM-style & 40.67 & 42.14 & 47.18 & 51.88 \\
Jev-DLC & \textbf{44.83} & \textbf{56.07} & \textbf{50.47} & \textbf{53.01} \\
\bottomrule
\end{tabular*}
\caption{TCPA-Repeat user-macro accuracy (\%). Every depth column evaluates
decisions made after at least $d$ earlier feedback events. Counts, paired
bootstrap intervals, and the complete baseline ladder appear in
Appendix~\appref{app:repeat-results}.}
\label{tab:repeat-main-results}
\end{table}

Persistent typed state improves TCPA-Repeat decisions above the frozen scorer.
Jev-DLC reaches 44.83\% on the full stream, a 5.39-point gain over Frozen and
a 2.43-point higher accuracy than Recent-4. It has the highest point estimate
among the reported external comparisons on Full and at each reported feedback
depth. After at least two earlier feedback events, it reaches 56.07\%,
7.54 points above the strongest reported comparison at this depth. Because
every prediction precedes its feedback, these differences measure the use of
earlier corrections rather than refitting the decisions that elicited them.

At $d\ge4$ and $d\ge8$, Jev-DLC reaches 50.47\% and 53.01\%, compared
with 47.18\% and 51.88\% for MRM-style, the strongest external reference in
these strata. Each stratum evaluates decisions with at least the specified
number of earlier feedback events. The deepest stratum contains 15 users;
Appendix~\appref{app:depth-results} gives the eligible counts and paired
intervals.

Appendix~\appref{app:repeat-results} reports sample sizes and paired
user-bootstrap intervals. The evaluation protocol is specified in
Appendix~\appref{app:reproducibility-scope}.

\subsection{Contribution of the Two Channels}

Retaining both correction trajectories improves later decisions over either
channel alone. Table~\ref{tab:channel-components} compares the count-informed
conservative channel (C), the directional responsive channel (R), and Jev-DLC
under the same frozen representation and channel parameters. Each variant
receives feedback from its own committed predictions.

\begin{table}[!ht]
\centering
\scriptsize
\setlength{\tabcolsep}{3.0pt}
\begin{tabular*}{\columnwidth}{@{\extracolsep{\fill}}lrrrr@{}}
\toprule
State & Full & $d\ge2$ & $d\ge4$ & $d\ge8$ \\
\midrule
C only & 43.12 & 54.85 & 48.60 & 44.30 \\
R only & 41.50 & 44.20 & 43.85 & 50.45 \\
Jev-DLC & \textbf{44.83} & \textbf{56.07} & \textbf{50.47} & \textbf{53.01} \\
\bottomrule
\end{tabular*}
\caption{Channel retention on TCPA-Repeat, in user-macro accuracy (\%).
Both channels contribute to the full Jev-DLC decision state.}
\label{tab:channel-components}
\end{table}

Jev-DLC reaches 44.83\% on Full, 1.71 points above C and 3.33 points above R.
The single channels have different depth profiles. C reaches 54.85\% after
at least two earlier feedback events, while R reaches 50.45\% in the deepest
stratum, compared with 44.30\% for C. Jev-DLC reaches 56.07\% and 53.01\%
in these two strata, respectively, and has the highest accuracy in every
reported column. The retained action counts and pairwise repair therefore
support complementary uses of earlier feedback within the same decision state.

The fusion analysis further compares adaptive weighting with equal and
development-fixed coefficients. Relative to the development-fixed combination,
adaptive Jev-DLC gains 0.68 points on Full and 4.86 points at $d\ge8$.
Appendix~\appref{app:fusion-strategies} reports the complete configurations
and paired intervals. Together, these comparisons distinguish the benefit of
retaining both channels from the benefit of adapting their relative weights.

\subsection{Serving State and Computation}

The methods reach later decisions through different serving objects.
Recent-4 preserves text-like evidence and expands the next scorer input.
PPlug stores a learned user vector, TAP-PER adds bounded vector memory, and MRM
optimizes a low-dimensional head. Jev-DLC stores two 32-dimensional correction
vectors, bounded semantic counts, and two reliability weights.

\begin{table}[t]
\centering
\scriptsize
\setlength{\tabcolsep}{2.6pt}
\begin{tabular*}{\columnwidth}{@{\extracolsep{\fill}}lrrll@{}}
\toprule
Method & State & Retained & Input & Runtime \\
\midrule
Recent-4 & 0 B & 1,166.5 B raw & +262.8 tok. & Longer fwd. \\
PPlug & 132 B & 0 B & 0 & Projection \\
TAP-PER & 128 B & 1,040 B vec. & 0 & Query \\
MRM & 16 B & 0 B & 0 & 3 low-D steps \\
Jev-DLC & 324 B & 0 B & 0 & Vector update \\
\bottomrule
\end{tabular*}
\caption{Maximum per-user serving objects on TCPA-Repeat. State excludes the
shared frozen scorer and train-time shared parameters.}
\label{tab:system-results}
\end{table}

The accuracy gain over Frozen comes from a bounded decision state, without
raw-history replay, an additional scorer forward, or runtime language-model
training. Its
324-byte maximum state is independent of stream length because the semantic
identities come from the fixed TCPA-Repeat ontology. This is a different
deployment tradeoff from replay, bounded vector memory, or low-dimensional
runtime optimization. Appendix~\appref{app:baseline-contracts} reports
latency, extra model computation, and complete state accounting.

After the frozen scorer output is available, state computation takes
0.1781 ms for prediction and 0.0810 ms for feedback updates on TCPA-Repeat.
These measurements cover CPU state operations.

Additional representation, accumulation, projection-size, seed, accept/revoke,
and direction controls appear in
Appendix~\appref{app:representation-analysis}. TCPA-Open and TCPA-Dynamics
results appear in Appendix~\appref{app:diagnostic-results}.

%% file: sections/08-conclusion.tex
\section{Conclusion}
\label{sec:conclusion}

A typed correction exposes information that can outlive the request that
elicited it. Jev-DLC preserves recurring action evidence and the
preferred--rejected relation in complementary bounded channels above a frozen
scorer. \benchmark{} makes this persistence measurable by crediting feedback
only through later decisions. On repeated actions, the resulting state improves
decision accuracy without retaining raw interactions or training the language
model at runtime. Together, the method and evaluation provide a decision-level
path to continual personalization in which user corrections remain useful
after their original requests have passed.

%% file: sections/09-limitations.tex
\section*{Limitations}

The current formulation uses immediate accept and revoke feedback over a finite
candidate set, so each correction maps directly to a decision relation. Future
work can recover this relation from delayed or free-form feedback, represent
richer signals such as ranked alternatives or partial acceptance, and place a
typed decision layer above open-ended generation. State migration when the
scorer or action ontology is upgraded is another useful direction, allowing
accumulated preferences to persist across system versions.

The repeated-action setting provides the clearest evidence for this state.
Changing-item recommendation favors PPlug overall.
The storage bound also depends on the evaluated semantic vocabularies; an
expanding action ontology requires a corresponding policy for count storage.

%% file: sections/appendix/00-contents.tex
\phantomsection
\begin{center}
{\Large Contents of Appendix}
\end{center}
\vspace{0.2cm}

\startcontents[appendix]
\printcontents[appendix]{l}{1}{\setcounter{tocdepth}{2}}

%% file: sections/appendix/a-notation-and-interface.tex
\section{Protocol, Notation, and Benchmark Construction}
\label{app:benchmark-protocol}

\subsection{Causal Serving Contract}
\label{app:causal-protocol}

Every track uses the same predict-then-update order. The runtime loads the
current public event and the user's state, commits a prediction, and only then
receives the evaluator-owned target as typed feedback. The feedback may update
the state for later events, but it cannot alter the decision that exposed it.
State is reset at user boundaries.

\begin{table}[t]
\centering
\small
\begin{tabular}{ll}
\toprule
Symbol & Meaning \\
\midrule
$x_t$ & current context \\
$\mathcal A_t$ & current candidate slate \\
$M_t(a)$ & immutable hard mask \\
$b_t(a)$ & frozen option logit \\
$h_t(a)$ & frozen candidate representation \\
$P$ & shared train-only linear projection \\
$r_t(a)$ & normalized projected representation \\
$n_t(k)$ & revealed count for semantic identity $k$ \\
$\mu_t^{\mathrm c},\mu_t^{\mathrm r}$ & conservative and responsive correction states \\
$g_t^{\mathrm c},g_t^{\mathrm r}$ & causal channel log weights \\
$\rho_t$ & responsive-channel mixture weight \\
$a_t^+,a_t^-$ & preferred and rejected typed actions \\
$d_t^j$ & channel-$j$ preferred--rejected contrast \\
$\beta_t^j$ & channel-$j$ frozen-plus-prior margin \\
$\gamma,C$ & margin and PA-I cap \\
\bottomrule
\end{tabular}
\caption{Notation shared by the frozen scorer, persistent state, and proofs.}
\label{tab:notation}
\end{table}

An accepted event sets $a_t^+=\hat a_t$ and uses the highest-scoring remaining
safe action as $a_t^-$. A revoked event sets $a_t^-=\hat a_t$ and
$a_t^+=a_t^\star$. Candidate identifiers must be unique within the event before
either action is dereferenced. An event without a valid typed correction causes
no state update.

\subsection{Runtime and Evaluator Boundary}
\label{app:object-boundary}

The implementation serializes model-visible events and evaluator-owned targets
separately. Table~\ref{tab:object-boundary} lists the fields crossing each
boundary. A runtime adapter can therefore be audited without access to target
files.

\begin{table*}[t]
\centering
\small
\setlength{\tabcolsep}{5pt}
\begin{tabular}{lll}
\toprule
Object & Fields & Visibility and role \\
\midrule
Candidate & candidate ID, type ID, text, metadata & visible before prediction \\
Runtime event & user ID, event ID, time, context, slate, track & visible before prediction \\
Prediction & event ID, choice, probability map, confidence, state version & committed before reveal \\
Target & event ID, target candidate/type, time, diagnostic labels & evaluator sidecar only \\
Feedback & event ID, kind, predicted ID, target ID, semantics & revealed after commit \\
State & counts, two correction vectors, reliability, version & persistent per user \\
\bottomrule
\end{tabular}
\caption{Serializable objects in the chronological protocol. Target and
diagnostic fields never enter the model-visible runtime event.}
\label{tab:object-boundary}
\end{table*}

\subsection{TCPA-Repeat Construction}
\label{app:repeat-construction}

TCPA-Repeat converts LaMP-2 into 15-way movie-tag decisions while preserving
the order of each user's source records. Users are disjoint across training,
development, and test. Candidate text and context remain in the runtime record,
while the selected tag appears only in the evaluator sidecar. Conversion is
deterministic and retains the source-record locator.

\begin{table}[t]
\centering
\small
\setlength{\tabcolsep}{5pt}
\begin{tabular}{lrrr}
\toprule
Split & Users & Events & Slate size \\
\midrule
Train & 747 & 5,216 & 15 \\
Development & 90 & 556 & 15 \\
Test & 92 & 711 & 15 \\
\bottomrule
\end{tabular}
\caption{TCPA-Repeat partitions after chronological user-disjoint
compilation.}
\label{tab:repeat-splits}
\end{table}

The stable action vocabulary creates repeated decision identities without
reordering the stream. The first event for each user has feedback depth zero.
At later events, depth counts only corrections revealed before the current
prediction. This differs from filtering users by total stream length, which is
reported separately when used.

\subsection{Additional Track Construction}
\label{app:diagnostic-construction}

TCPA-Open is derived from MIND impression logs. Each event presents a four-item
slate sampled within one impression, and each later impression introduces a
new set of items. Its train, development, and test partitions contain
79,849, 10,221, and 9,845 events; the test partition contains 4,209 users.
This track tests whether feedback transfers through representation rather than
through a recurring action identity.

TCPA-Dynamics contains stable, partial-shift, recurrence, and
20\%-label-noise streams. Each test regime contains eight users and
512 evaluated events. Partial shift changes only part of the latent
preference, recurrence restores its initial value after a shifted phase, and
the evaluator sidecar retains the clean target and phase labels required for
recovery, retention, obsolete-persistence, and recurrence metrics.

\subsection{Validation and Leakage Audits}
\label{app:leakage-audit}

The validator rejects duplicate event IDs, targets outside the current slate,
runtime/evaluator mismatches, and target-like fields in model-visible records.
User grouping precedes splitting. Cache exporters read runtime files only and
record the runtime, prompt, model, representation, and output hashes. The
runner scores the committed prediction before reading feedback.

Each frozen result records the source locator, candidate-order hash, runtime
hash, evaluator hash, model revision, representation revision, and output hash.
These fields distinguish target leakage, row mismatch, and accidental reuse of
a result under another method key. The runner also checks that each prediction
contains one normalized probability per safe candidate, user streams are
ordered before replay, and the state version advances only after feedback.

\subsection{Metrics and Statistical Inference}
\label{app:metrics}

User-macro accuracy averages within user before averaging across users.
Event-micro accuracy pools events. NLL, ten-bin ECE, and the slate-normalized
multiclass Brier score are secondary metrics. Confidence intervals use 10,000
paired bootstrap draws over users. Each draw samples user IDs with replacement
and includes both methods' complete records for every sampled user.

For feedback-depth subsets, eligibility is determined from the common stream
before either method is scored. A prediction at depth $k$ has exactly $k$
earlier feedback events available. This causal definition prevents current or
future targets from deciding whether an event enters a reported stratum.

%% file: sections/appendix/c-method-details.tex
\section{Model, Representation, and Baseline Details}
\label{app:method-details}

\subsection{Typed Interface and Scorer Implementation}
\label{app:interface-implementation}

Jev serves as a typed decision model in our setting. Given the current context
and a finite set of candidate actions, it selects an action and returns its
identifier. Acceptance confirms that choice; revocation names the preferred
replacement. This interaction exposes the decision relation that Jev-DLC
retains for later requests.

The complete DLC implementation uses local JEV-SFT, trained offline for typed
decisions. It exposes candidate logits and hidden representations for the
shared projection and two-channel state. The scorer remains frozen while
revealed feedback updates the per-user state.

The hosted Jev runs evaluate current-event decision quality without
replaying earlier TCPA interactions or maintaining continual adaptation state. They cover all 711 Repeat and 9,845 Open events
and provide frozen anchors. Local JEV-SFT supplies the representations used
by the complete DLC experiments. The hosted run and prompt contract are
specified in Appendix~\ref{app:prompts-api-cases}.

\subsection{Frozen Scorer Training}
\label{app:scorer-training}

\jone{} adapts Qwen3.8-27B with LoRA rank 8 and scale 16. The prompt ends
before the answer token, and the objective normalizes only the candidate letter
logits,
\[
 \mathcal L_{\mathrm{SFT}}=-\log
 \frac{\exp b_t(a_t^\star)}
 {\sum_{a\in\mathcal A_t}\exp b_t(a)}.
\]
Training alternates balanced TCPA-Repeat and TCPA-Open records for 300
optimizer steps, with gradient accumulation 4, learning rate $10^{-4}$, and
maximum length 2,048. Training, development, and test users are disjoint.

\subsection{Learned Candidate Projection}
\label{app:learned-projection}

For every candidate, the target-free exporter records the base option logit and
the final-layer hidden vector at the candidate's option line. The hidden
dimension is 5,120. Runtime records, model and adapter revisions, event order,
candidate counts, and cache hashes are checked before the representation is
used. The exporter reads no evaluator fields.

The shared projection $P\in\mathbb R^{32\times5120}$ is fitted on training
users only. Training replays each user chronologically. At event $t$, the loss
scores the current target using state formed from events before $t$, then uses
the revealed training feedback to update that state for the next event. We use
six AdamW epochs, learning rate .03, weight decay $10^{-4}$, gradient clipping
at 5, and seed 20260927. Projection rows are normalized after each epoch.
Development and test users never update $P$.

At serving time, each hidden vector is projected and unit normalized. The
projection is shared model state rather than personalized storage. Both
Jev-DLC channels use this same frozen representation.

\subsection{Jev-DLC Configuration}
\label{app:jev-dlc-configuration}

Both correction vectors and all semantic counts begin at zero. The conservative
channel uses Dirichlet smoothing $\alpha=0.5$, count weight $2.0$, and
representation weight $\lambda_p=2.0$; the responsive channel uses the
unscaled representation. Both PA-I repairs use margin $\gamma=1$ and cap
$C=1$. Revocation uses the explicit withdrawn-to-corrected pair. Acceptance
contrasts the accepted decision with its highest-probability remaining safe
action.

The conservative log weight begins at $2.0$ and the responsive log weight at
zero. Channel weighting uses zero-one loss, learning rate $\eta=.25$, geometric
probability mixing, and fixed share $\sigma=.1$. Each prediction is committed
before feedback updates the counts, correction vectors, or channel weights.
The same configuration is used on TCPA-Repeat and TCPA-Open. The evaluated
maximum state occupies 324 bytes on Repeat and 320 bytes on Open, including
two float32 log weights. The count-state bound depends on the evaluated
semantic vocabulary. Appendix~\ref{app:reproducibility-scope} records the
evaluation protocol; Appendix~\ref{app:fusion-configuration} specifies
the other fusion configurations.

\subsection{Matched Baseline Contracts}
\label{app:baseline-contracts}

All methods share event identity, safe slates, revealed targets, user grouping,
and chronological evaluation. PPlug, TAP-PER, MRM, and Jev-DLC use the learned
projection of the frozen JEV-SFT cache. Recent-4 reruns the same checkpoint
with a history-augmented prompt and therefore does not share identical cached
representations. The archived AdaGrad and older online ladder are retained
without refitting; their representation provenance requires reconciliation
before calling them learned-projection-matched controls. Shared trainable
components use training users, and the fusion configurations are
specified separately below.

\paragraph{Recent-4.}
Recent-4 retains the four most recent typed interactions, reconstructs the
history block, and reruns JEV-SFT on the enlarged prompt. It represents the
history-replay alternative in which stored evidence is reinterpreted for every
new decision.

\paragraph{AdaGrad and the complete online ladder.}
AdaGrad stores a 32-dimensional state and a diagonal squared-gradient
accumulator. ONS, DER++-8, and Directional Drift extend this comparison with
full-matrix geometry, bounded vector replay, and covariance-aware drift,
respectively. These archived rows use causal pairwise feedback; their
representation provenance is qualified above. Complete results appear in
Appendix~\ref{app:baseline-depth-results}.

\paragraph{PPlug.}
PPlug learns a compact representation of user behavior rather than replaying a
profile at inference time \citep{liu2025pplug}. Our matched instantiation fits a
shared $32\times32$ projection on training users and applies it to the causal
running mean of preferred candidate representations. The per-user object is the
32-dimensional sum and its count; no test feedback updates the shared map.

\paragraph{TAP-PER.}
TAP-PER separates a persistent user representation from a temporally weighted,
query-conditioned component \citep{cao2026tapper}. The matched TCPA version
maintains a global 32-dimensional correction state and a bounded memory of eight
slate-key and correction-vector pairs. The current slate queries this memory
with development-selected temporal decay and temperature before the next
decision is committed.

\paragraph{MRM.}
MRM represents a personalized reward model as a low-dimensional combination of
shared reward functions and adapts the coefficients from limited user feedback
\citep{cai2026mrm}. Our matched instantiation learns a shared rank-four basis on
training users and retains four coefficients per test user. Each revealed typed
correction performs three gradient steps only in this four-dimensional space;
the frozen scorer and shared basis remain unchanged.

\begin{table*}[t]
\centering
\small
\setlength{\tabcolsep}{3.5pt}
\begin{tabular}{lllllr}
\toprule
Method & Per-user state & Retained history & Extra input & Runtime update & Bytes \\
\midrule
Frozen & none & none & 0 & none & 0 \\
Recent-4 & none & 4 typed interactions & +262.8 tok./event & scorer forward & 1,166.5 mean \\
AdaGrad & state + diagonal accumulator & none & 0 & vector update & 256 \\
PPlug & sum + count & none & 0 & projection & 132 \\
TAP-PER & global state & 8 vector pairs & 0 & memory query & 1,168 \\
MRM & 4 coefficients & none & 0 & 3 low-D steps & 16 \\
\method{} & two states + counts + reliability & none & 0 & vector update & 324 \\
\bottomrule
\end{tabular}
\caption{Matched serving contracts on TCPA-Repeat. Byte counts exclude the
shared frozen scorer, projection, and train-time shared parameters. TAP-PER's
total includes its 128-byte state and at most 1,040 bytes of vector memory.}
\label{tab:baseline-accounting}
\end{table*}

\subsection{Timing Protocol}
\label{app:timing-protocol}

State latency is averaged after the shared scorer output exists. On
TCPA-Repeat, Jev-DLC uses 0.1781 ms for prediction-side state
computation and 0.0810 ms for the feedback update in the current CPU replay. Recent-4 and the $K=0$ reference are
measured end to end on the same server, checkpoint, adapter, batch size 4,
maximum length 4,096, and test stream. CUDA synchronization brackets each
forward. The $K=0$ forward takes 300.54 ms per event and Recent-4 takes
449.90 ms.

Update-only timing and a full model forward measure different work. The system
comparison therefore reports retained bytes, added input tokens, scorer calls,
and backward passes alongside latency rather than collapsing them into one
number.

\subsection{Other Fusion Configurations}
\label{app:fusion-configuration}

Fixed fusion uses one responsive coefficient throughout each replay and
disables the complete channel-weight transition, including fixed share.
Equal fusion uses $\rho=.5$; development-fixed fusion selects its coefficient
on development users before test evaluation. All variants retain the frozen
representation and channel parameters and receive feedback from their own
committed predictions. Appendix~\ref{app:fusion-strategies} reports the
collaborator experiment comparing these configurations.

%% file: sections/appendix/b-theory-and-proofs.tex
\section{Theory and Proofs}
\label{app:theory}

\subsection{Closed-Form PA-I Solution}

For either channel $j$, abbreviate $d=d_t^j$, $\beta=\beta_t^j$,
$\mu_0=\mu_t^j$, and $\ell=[\gamma-\beta-\mu_0^\top d]_+$. The primal problem is
\[
 \min_{\mu,\xi\ge0}\tfrac12\lVert\mu-\mu_0\rVert^2+C\xi
 \quad\text{s.t.}\quad \gamma-\xi-\beta-\mu^\top d\le0.
\]

\paragraph{Local repair proposition.}
If $d\ne0$, the unique state solution is
$\mu^\star=\mu_0+\tau d$ with
$\tau=\min(C,\ell/\lVert d\rVert^2)$. If $d=0$, the implementation leaves the
state unchanged and absorbs any violation in slack.

\paragraph{Proof.}
Introduce multipliers $\tau\ge0$ for the margin constraint and $\lambda\ge0$
for $\xi\ge0$. The Lagrangian is
\[
 L=\tfrac12\lVert\mu-\mu_0\rVert^2+C\xi
 +\tau(\gamma-\xi-\beta-\mu^\top d)-\lambda\xi.
\]
Stationarity gives $\mu-\mu_0=\tau d$ and
$C-\tau-\lambda=0$, hence $0\le\tau\le C$. Substitution into the dual leaves
the concave quadratic
\[
 \max_{0\le\tau\le C}\quad \tau\ell-\tfrac12\tau^2\lVert d\rVert^2.
\]
Its maximizer is the stated clipped ratio. The objective is strictly convex in
$\mu$, so the state component is unique. When $\ell=0$, the maximizer is zero
and the update is passive. \hfill$\square$

\subsection{KKT Cases and Degenerate Events}

The clipped ratio separates the update into three cases. If $\ell=0$, the
current state already lies in the feasible half-space and the unique minimizer
is $\mu_0$. If $0<\ell<C\lVert d\rVert^2$, the cap is inactive, the margin
constraint is tight, and the solution is the orthogonal projection onto its
boundary. If $\ell\ge C\lVert d\rVert^2$, the dual variable reaches $C$ and the
remaining violation is absorbed by positive slack. Thus $C$ limits the
influence of one event without changing the update direction.

When $d=0$, the feedback pair is indistinguishable under the frozen
representation. No state in this representation can alter its pairwise score,
so the implementation records the event but leaves $\mu$ unchanged. This is a
representation boundary rather than an optimization failure. If the current pairwise margin $\beta_t+\mu_t^\top d_t$ already meets
the required margin $\gamma$, the feedback leaves the correction vector
unchanged.

\subsection{Bounded Movement and Minimality}

\paragraph{Bounded-step corollary.}
If $\lVert x_t^j(a)\rVert\le\kappa_j$ for every candidate, then each channel
transition obeys
$\lVert\mu_{t+1}^j-\mu_t^j\rVert\le2C\kappa_j$.

\paragraph{Proof.}
The triangle inequality gives $\lVert d_t^j\rVert\le2\kappa_j$, while the
local repair proposition gives $\tau_t^j\le C$. Thus
$\lVert\mu_{t+1}^j-\mu_t^j\rVert
=\tau_t^j\lVert d_t^j\rVert\le2C\kappa_j$.
Each channel remains fixed-dimensional independent of stream length.
\hfill$\square$

When $C$ is not active, the solution is the Euclidean projection of $\mu_t$
onto the half-space $\{\mu:\beta_t+\mu^\top d_t\ge\gamma\}$. Consequently,
every zero-slack state satisfying the new constraint lies at least
$\ell_t/\lVert d_t\rVert$ from $\mu_t$, and \method{} attains this distance.
This is the precise minimum-change statement; with an active cap, PA-I instead
chooses the optimum of the stated slack-penalized problem.

\subsection{Comparator-Relative Cumulative Bound}

Define $\ell_t(u)=[\gamma-\beta_t-u^\top d_t]_+$ for any fixed comparator
$u\in\mathbb R^d$.

\paragraph{Comparator-relative violation bound.}
For $\mu_1=0$ and the PA-I trajectory,
\[
 \begin{aligned}
 2\sum_{t=1}^T\tau_t\ell_t
 &-\sum_{t=1}^T\tau_t^2\lVert d_t\rVert^2 \\
 &\le \lVert u\rVert^2+2C\sum_{t=1}^T\ell_t(u).
 \end{aligned}
\]

\paragraph{Proof.}
Only rounds with $\ell_t>0$ move. On such a round,
$\mu_t^\top d_t=\gamma-\beta_t-\ell_t$. Since
$u^\top d_t\ge\gamma-\beta_t-\ell_t(u)$,
\[
 d_t^\top(\mu_t-u)\le-\ell_t+\ell_t(u).
\]
Expanding the squared distance after
$\mu_{t+1}=\mu_t+\tau_t d_t$ yields
\[
 \begin{aligned}
 \lVert\mu_{t+1}-u\rVert^2
 &\le \lVert\mu_t-u\rVert^2 \\
 &\quad-2\tau_t\ell_t+2\tau_t\ell_t(u) \\
 &\quad+\tau_t^2\lVert d_t\rVert^2.
 \end{aligned}
\]
Rearrange, sum over $t$, use the nonnegativity of the final squared distance,
$\mu_1=0$, and $\tau_t\le C$. This gives the result. \hfill$\square$

If one comparator satisfies all constraints and the cap is inactive, then
$\ell_t(u)=0$ and $\tau_t=\ell_t/\lVert d_t\rVert^2$. Substitution gives
\[
 \sum_t\frac{\ell_t^2}{\lVert d_t\rVert^2}\le\lVert u\rVert^2.
\]
The theorem is comparator-relative. The controlled shift and recurrence tracks
complement it by measuring how fixed-state updates behave as preferences move.

\subsection{Action Sufficiency and Realization Nuisance}

\paragraph{Action-sufficiency proposition.}
Let realization $R$ map deterministically to action $A=g_x(R)$. If
$p_{w,\zeta}(R\mid x)=q_\zeta(R\mid A,x)p_w(A\mid x)$ and $q_\zeta$ is
independent of $w$, then $A$ is sufficient for $w$ conditional on $x$.

\paragraph{Proof.}
The likelihood factors into a term independent of $w$ and a term depending on
$R$ only through $A$. The Neyman--Fisher factorization criterion gives the
claim. Products of conditionally independent events preserve the factorization.
\hfill$\square$

\paragraph{Typed Rao--Blackwellization.}
For any square-integrable update $V(R,F)$ and
$\bar V(A,F)=\mathbb E[V\mid A,F]$,
\[
 \operatorname{Cov}(V)=\operatorname{Cov}(\bar V)
 +\mathbb E[\operatorname{Cov}(V\mid A,F)].
\]
Hence $\operatorname{Cov}(\bar V)\preceq\operatorname{Cov}(V)$, with strict
trace reduction exactly when realization-dependent conditional variance is
nonzero.

\paragraph{Proof.}
The tower property gives equal expectations. Write
$V-\mathbb EV=(V-\bar V)+(\bar V-\mathbb EV)$. The cross terms vanish because
$\mathbb E[V-\bar V\mid A,F]=0$. Taking outer products and expectations gives
the covariance decomposition. The remaining conditional covariance is
positive semidefinite, and its trace is
$\mathbb E\lVert V-\bar V\rVert^2$. \hfill$\square$

These statements define an equality boundary. If a token-level update already
depends on text only through the same action identity, typed conditioning does
not improve its statistical efficiency.

For a conditionally independent sequence of realizations, the factorization
extends by multiplication. The typed action sequence is then sufficient for
the preference parameter given the context sequence. This sequence statement
justifies accumulating decision-level state without asserting that arbitrary
free-form feedback can be reduced to actions. If the realization mechanism
itself depends on $w$, or if two surface forms mapped to the same action carry
different preference information, the sufficient-statistic argument no longer
applies.

\subsection{Tracking Channel Reliability}
\label{app:channel-weighting-theory}

Let $w_t^j$ be the pre-decision weight of channel
$j\in\{\mathrm c,\mathrm r\}$ and let $L_t^j\in[0,1]$ be its revealed loss.
The fixed-share transition is
\[
 Q(j'\mid j)=(1-\sigma)\mathbf 1[j'=j]+\frac{\sigma}{2}.
\]
For a channel path $j_{1:T}$ with $m$ switches, its prior mass is
\[
 \Pi(j_{1:T})=\pi_1^{j_1}
 \prod_{t=1}^{T-1}Q(j_{t+1}\mid j_t).
\]

\paragraph{Fixed-share tracking theorem.}
For $0<\sigma<1$, define
$\mathcal R_T=\sum_{t=1}^T(\sum_jw_t^jL_t^j-L_t^{j_t})$.
The reliability update satisfies
\[
\begin{aligned}
 \mathcal R_T
 &\le \frac{-\log\pi_1^{j_1}
 -(T-1-m)\log(1-\sigma/2)}{\eta}\\
 &\quad+\frac{-m\log(\sigma/2)}{\eta}+\frac{\eta T}{8}.
\end{aligned}
\]

\paragraph{Proof.}
Regard each channel path as an expert with prior mass $\Pi(j_{1:T})$.
Exponential weighting over these path experts produces the same channel
marginals as the fixed-share recursion, because multiplying by $Q$ after each
loss update exactly redistributes the posterior mass between channels. The
standard exponential-weights potential argument and Hoeffding's lemma give
\[
 \sum_t\sum_j w_t^jL_t^j-\sum_tL_t^{j_t}
 \le \frac{-\log\Pi(j_{1:T})}{\eta}+\frac{\eta T}{8}.
\]
The path contains $m$ transitions of probability $\sigma/2$ and
$T-1-m$ stay transitions of probability $1-\sigma/2$. Substituting their
product yields the stated result. \hfill$\square$

The theorem bounds the weighted channel loss $\sum_jw_t^jL_t^j$ relative
to a switching channel path. This quantity is distinct from the zero-one loss
of the deterministic argmax of the geometrically mixed distribution. The
bound describes the weighting recursion and does not establish an accuracy
advantage over another fusion implementation.

\subsection{Mask and Reset Invariance}

Masked actions are excluded before either channel softmax. They therefore
retain probability zero after geometric mixing. Reset clears semantic counts,
sets both correction vectors to zero, and returns the reliability weights to
their initialization. With the preference components disabled, the serving
path restores the frozen distribution over the unchanged safe set. These
properties follow from the serving order and do not depend on the observed
feedback sequence.

%% file: sections/appendix/d-benchmark-and-settings.tex
\section{TCPA-Repeat Results and System Comparisons}
\label{app:repeat-results}

\subsection{Frozen Scorer Anchors}
\label{app:scorer-anchors}

The scorer comparison fixes current-event decision quality before any
adaptation. Raw Qwen ranks the candidate option tokens with the unadapted base
model. \jone{} uses the frozen local checkpoint described in
Appendix~\ref{app:scorer-training}. Hosted Jev receives the same current event
through its typed choice API and receives no user history.

\begin{table}[t]
\centering
\small
\begin{tabular}{lr}
\toprule
Frozen scorer & User-macro accuracy (\%) \\
\midrule
Raw Qwen option logprob & 26.93 \\
\jone{} & 39.44 \\
Hosted Jev & 41.22 \\
\bottomrule
\end{tabular}
\caption{Current-event scorer anchors on the TCPA-Repeat test stream.}
\label{tab:repeat-scorer-anchors}
\end{table}

\subsection{Feedback-Depth Results}
\label{app:depth-results}

Table~\ref{tab:repeat-depth-complete} reports the full causal evaluation behind
the main accuracy table. Depth is measured immediately before the current
prediction. Every confidence interval is obtained by paired resampling of the
eligible users.

\begin{table}[t]
\centering
\scriptsize
\setlength{\tabcolsep}{2.1pt}
\begin{tabular}{lrrrrrl}
\toprule
Prior depth & Users & Events & Frozen & Jev-DLC & Gain & 95\% CI \\
\midrule
$\ge0$ & 92 & 711 & 39.44 & 44.83 & +5.39 & $[+2.04,+8.74]$ \\
$\ge1$ & 85 & 619 & 39.21 & 46.67 & +7.46 & $[+2.12,+12.76]$ \\
$\ge2$ & 63 & 534 & 41.58 & 56.07 & +14.49 & $[+7.42,+22.63]$ \\
$\ge4$ & 29 & 434 & 36.48 & 50.47 & +13.99 & $[+2.01,+27.42]$ \\
$\ge8$ & 15 & 339 & 40.38 & 53.01 & +12.63 & $[+2.31,+27.23]$ \\
\bottomrule
\end{tabular}
\caption{TCPA-Repeat user-macro accuracy for Jev-DLC. Gains and
paired user-bootstrap intervals are in percentage points.}
\label{tab:repeat-depth-complete}
\end{table}

The number of eligible users decreases with depth, but the point estimates and
paired intervals preserve the same causal unit. The depth-$\ge8$ result uses
339 decisions from 15 users, rather than selecting users by their eventual
labels or method outcomes.

\subsection{Fixed-Cohort Depth Check}
\label{app:fixed-cohort-depth}

The archived directional PA-I diagnostic uses the same 15 users at every
reported depth threshold. It measures how the local correction state behaves
under a fixed cohort, separately from the C/R fusion experiment. Its
within-user slope is $+0.71$ points per prior feedback, with a 95\% bootstrap
interval of $[-0.34,+2.04]$. The main depth tables report accuracy on events
with at least the stated amount of earlier feedback; their eligible cohorts
change with the threshold.

\begin{table}[t]
\centering
\small
\setlength{\tabcolsep}{3.2pt}
\begin{tabular}{lrrr}
\toprule
Prior depth & Events & Gain & 95\% CI \\
\midrule
$\ge0$ & 459 & +7.08 & $[+3.18,+12.18]$ \\
$\ge1$ & 444 & +7.57 & $[+3.35,+13.14]$ \\
$\ge2$ & 429 & +7.65 & $[+2.92,+13.90]$ \\
$\ge4$ & 399 & +8.13 & $[+2.31,+16.24]$ \\
$\ge8$ & 339 & +12.20 & $[+2.54,+26.75]$ \\
\bottomrule
\end{tabular}
\caption{Fixed-cohort depth diagnostic for the archived directional PA-I replay.
All rows use the same 15 users; gains are in percentage points over Frozen.}
\label{tab:repeat-fixed-cohort}
\end{table}

\subsection{Representative Baseline Ladder}
\label{app:baseline-depth-results}

The complete depth matrix in Table~\ref{tab:repeat-baseline-depth} supplements
the compact main-text table. Every cell reports absolute user-macro accuracy on
the common eligible events. Representation and mutable state for each row are
given in Table~\ref{tab:baseline-accounting}.

\begin{table}[t]
\centering
\small
\setlength{\tabcolsep}{3.4pt}
\begin{tabular}{lrrrr}
\toprule
Method & Full & $d\ge2$ & $d\ge4$ & $d\ge8$ \\
\midrule
Frozen & 39.44 & 41.58 & 36.48 & 40.38 \\
Recent-4 & 42.40 & 46.40 & 37.56 & 42.17 \\
AdaGrad & 41.99 & 48.53 & 39.33 & 41.68 \\
DER++-8 & 41.94 & 48.45 & 39.14 & 41.20 \\
Directional Drift & 41.76 & 43.26 & 38.61 & 44.71 \\
ONS & 40.70 & 42.14 & 37.85 & 42.01 \\
PPlug & 40.37 & 41.77 & 37.86 & 43.38 \\
TAP-PER & 41.99 & 46.48 & 46.45 & 44.25 \\
MRM & 40.67 & 42.14 & 47.18 & 51.88 \\

\method{} & \textbf{44.83} & \textbf{56.07} & \textbf{50.47} & \textbf{53.01} \\
\bottomrule
\end{tabular}
\caption{TCPA-Repeat user-macro accuracy across prior feedback depth. The
table reports absolute user-macro accuracy at each prior-feedback depth.
Other fusion configurations are reported separately in
Table~\ref{tab:fusion-strategies}.}
\label{tab:repeat-baseline-depth}
\end{table}

\subsection{Action-Frequency Variants}
\label{app:action-prior}

The action-frequency variants provide a simple alternative to typed pairwise
state. Both consume only feedback revealed before the current event and never
inspect the current target. The count prior adds a smoothed per-user count for
each action. The recency prior uses the same count signal with exponential
decay. Development selection gives the count prior $\alpha=0.1$ and weight
$1.0$; the recency prior uses $\alpha=0.1$, weight $1.0$, and decay $0.97$.

\begin{table}[t]
\centering
\scriptsize
\setlength{\tabcolsep}{2.6pt}
\begin{tabular}{lrrrrr}
\toprule
Method & Full & $d\ge1$ & $d\ge2$ & $d\ge4$ & $d\ge8$ \\
\midrule
Frozen & 39.44 & 0.00 & 0.00 & 0.00 & 0.00 \\
Action count prior & 43.74 & +5.69 & +12.76 & +13.23 & +6.08 \\
Recency action prior & 43.87 & +5.86 & +13.01 & +14.07 & +6.05 \\

\method{} & 44.83 & +7.46 & +14.49 & +13.99 & +12.63 \\
\bottomrule
\end{tabular}
\caption{Causal action-frequency variants on TCPA-Repeat. Full reports
user-macro accuracy in percent; depth columns report gain over Frozen in
percentage points.}
\label{tab:action-prior-depth}
\end{table}

Action counts retain which identities recur. Jev-DLC additionally preserves
the preferred--rejected relation, so a correction can change the ordering of
later candidates according to their representation alignment. The two-channel
design retains both forms of evidence instead of reconstructing them from
interaction text.

Table~\ref{tab:action-prior-depth} reports the separately retained local
frequency-control replay. Relative to the recency prior, Jev-DLC gains
$+0.96$ points on Full and $+6.58$ at $d\ge8$; the paired 95\% intervals are
$[-1.29,+3.43]$ and $[-0.57,+20.16]$, respectively.
Table~\ref{tab:fusion-strategies} evaluates the contribution of each channel
and their combination.

\subsection{History Replay and Serving Cost}
\label{app:recent-k}

Recent-$K$ serializes only interactions revealed before the current event.
An accept record names the accepted candidate, while a revoke record names both
the withdrawn and corrected candidates. Development selects $K=4$ over $K=8$
by user-macro accuracy, then NLL, then smaller $K$.

\begin{table}[t]
\centering
\small
\setlength{\tabcolsep}{3.2pt}
\begin{tabular}{lrrrrr}
\toprule
Split/$K$ & Acc. & NLL & Extra tok. & Hist. B & Forward ms \\
\midrule
Dev/4 & 48.20 & 1.549 & 241.5 & 1,082.2 & 453.10 \\
Dev/8 & 47.89 & 1.547 & 392.8 & 1,759.7 & 296.75 \\
Test/0 & 38.98 & 1.963 & 0 & 0 & 300.54 \\
Test/4 & 42.40 & 1.791 & 262.8 & 1,166.5 & 449.90 \\
\bottomrule
\end{tabular}
\caption{Prompt reconstruction with the frozen scorer. The matched test runs
provide the serving comparison; development latency is descriptive because
$K=4$ and $K=8$ ran sequentially. Test/0 reports a separate empty-history forward
run of the same checkpoint for matched latency measurement; Frozen reports
predictions from the scorer cache used for online adaptation.}
\label{tab:recent-k-full}
\end{table}

At event $t$, the history block contains at most the previous $K$ feedback
records and never the current target. The byte and token means include early
events whose available histories are shorter than four records. The fresh
$K=0$ forward is used only as the latency anchor; frozen-cache accuracy remains
the common accuracy reference for online methods.

\begin{table}[t]
\centering
\scriptsize
\setlength{\tabcolsep}{2pt}
\begin{tabular}{lrrrr}
\toprule
Method & Acc. & Retained & Extra tokens & Backward \\
\midrule
Frozen & 39.44 & 0 B & 0 & No \\
\method{} & 44.83 & 324 B state & 0 & No \\
Recent-4 & 42.40 & 1,166.5 B history & 262.8/event & No \\
\bottomrule
\end{tabular}
\caption{TCPA-Repeat serving footprint. Jev-DLC uses 0.1781 ms for
prediction-side state computation and 0.0810 ms for the feedback update;
Recent-4 reruns the scorer over its enlarged prompt.}
\label{tab:repeat-system-complete}
\end{table}

\subsection{Chronological Linear Reference}
\label{app:causal-headroom}

The chronological reference fits a per-user linear head only on the prefix available
before each event. It never uses the current or future target for scoring. The
reference is diagnostic rather than a deployable baseline because it repeatedly
fits a user-specific model.

\begin{table}[t]
\centering
\small
\setlength{\tabcolsep}{4pt}
\begin{tabular}{lrrr}
\toprule
TCPA-Repeat stratum & Frozen & Prefix-fit ref. & Gain \\
\midrule
Full stream & 39.44 & 42.56 & +3.11 \\
Length $\ge4$ & 43.56 & 49.51 & +5.94 \\
Length $\ge8$ & 48.56 & 54.86 & +6.30 \\
\bottomrule
\end{tabular}
\caption{Chronological personalization headroom in user-macro accuracy
percentage points. Stream length is distinct from prior feedback depth.}
\label{tab:strict-causal-headroom}
\end{table}

\subsection{Paired External Comparisons}
\label{app:paired-external}

The paired bootstrap resamples users with replacement for 10,000 draws using
seed 20260928. The following intervals compare the main Jev-DLC implementation
with every external row on the full Repeat stream. These marginal intervals
are reported without a multiplicity adjustment.

\begin{table}[t]
\centering
\small
\begin{tabular}{lrl}
\toprule
Comparison & Difference & 95\% CI \\
\midrule
Recent-4 & +2.43 & $[-1.33,+6.30]$ \\
Archived AdaGrad & +2.84 & $[-0.60,+6.20]$ \\
PPlug & +4.47 & $[+1.15,+7.71]$ \\
TAP-PER & +2.84 & $[-0.39,+6.13]$ \\
MRM & +4.16 & $[+0.64,+7.71]$ \\
\bottomrule
\end{tabular}
\caption{Main Jev-DLC minus each archived external comparison on
TCPA-Repeat Full, in percentage points.}
\label{tab:paired-external-fixed}
\end{table}

%% file: sections/appendix/e-additional-results.tex
\section{Representation and Mechanism Analyses}
\label{app:representation-analysis}

\subsection{Directional Representation and Accumulation Controls}
\label{app:representation-ablation}

Both Jev-DLC channels use the train-only learned linear projection described
in Appendix~\ref{app:learned-projection}. Separately archived directional
PA-I controls compare this map with a one-hidden-layer MLP of width 64 and
compare PA-I with running-mean accumulation. These controls isolate
representation and update choices; C/R retention is evaluated in
Table~\ref{tab:fusion-strategies}. Every learned encoder is shared across
users, fitted on training users, frozen at test time, and outputs a
32-dimensional representation.

\begin{table}[t]
\centering
\scriptsize
\setlength{\tabcolsep}{2.5pt}
\begin{tabular}{llrr}
\toprule
Representation & Update & TCPA-Repeat & TCPA-Open \\
\midrule
Frozen scorer & none & 39.44 & 28.04 \\
Learned linear & Mean & 40.50 & 28.63 \\
Learned linear & PA-I & 41.95 & 28.70 \\
MLP64 & Mean & 43.05 & 28.69 \\
MLP64 & PA-I & 42.50 & 28.78 \\
\bottomrule
\end{tabular}
\caption{User-macro test accuracy for archived directional representation and
accumulation controls. These rows vary the local correction path separately
from the two-channel fusion experiment.}
\label{tab:representation-update-ablation}
\end{table}

In the directional controls, MLP64 PA differs from learned linear PA by $+0.55$ points on TCPA-Repeat and
$+0.08$ on TCPA-Open. The corresponding paired 95\% intervals,
$[-1.02,+2.32]$ and $[-0.18,+0.33]$, include zero.

\subsection{Depth Behavior across Representations}
\label{app:representation-depth}

\begin{table}[t]
\centering
\scriptsize
\setlength{\tabcolsep}{2pt}
\begin{tabular}{lrrrr}
\toprule
Projection with PA-I & Full & $d\ge1$ & $d\ge4$ & $d\ge8$ \\
\midrule
Learned linear & +2.50 & +3.76 & +8.51 & +12.20 \\
MLP64 & +3.06 & +5.04 & +10.08 & +10.85 \\
\bottomrule
\end{tabular}
\caption{TCPA-Repeat gain over Frozen in percentage points.}
\label{tab:representation-depth}
\end{table}

MLP64 has the larger point estimate on the full stream and at depth $\ge4$,
while the learned linear projection reaches the larger gain at depth $\ge8$.
The two representations therefore show different depth profiles without an
established overall advantage. The full method uses the learned linear map
shared by both channels; MLP64 remains a nonlinear diagnostic.

\subsection{Future-Predictive Geometry}
\label{app:representation-geometry}

For correction direction $d_{u,t}$, the past-centroid diagnostic compares the
mean of $d_{u,1:t-1}$ with $d_{u,t}$. The matched-state diagnostic compares the
state built from the same user's past with the future direction; the shuffled
control replaces it with another user's state. These cosines diagnose whether
earlier corrections align with later pairs, rather than identifying a global
user direction.

\begin{table}[t]
\centering
\scriptsize
\setlength{\tabcolsep}{2pt}
\begin{tabular}{lrrr}
\toprule
Track & Past centroid & Matched & Shuffled \\
\midrule
TCPA-Repeat & .340 & .285 & -.046 \\
TCPA-Open & -.027 & -.027 & .005 \\
\bottomrule
\end{tabular}
\caption{Cosine diagnostics between past state and future correction
directions. TCPA-Repeat exhibits alignment-conditioned reuse; TCPA-Open does
not show the same geometry.}
\label{tab:representation-geometry}
\end{table}

\subsection{Jev-DLC Channel and Fusion Configurations}
\label{app:fusion-strategies}

Jev-DLC preserves both recurrence evidence and correction geometry. Here, we
systematically evaluate alternative channel retention and score-combination
strategies while holding the frozen representation and channel parameters
fixed. Throughout this analysis, $C$ and $R$ denote the conservative and
responsive channels, respectively. Each variant traces its own committed-feedback
trajectory under identical causal evaluation protocols.
Appendix~\ref{app:fusion-configuration} describes the fixed-weight variants
and their development-based selection.

\begin{table}[t]
\centering
\scriptsize
\setlength{\tabcolsep}{3.5pt}
\begin{tabular}{lrrrr}
\toprule
Configuration & Full & $d\ge2$ & $d\ge4$ & $d\ge8$ \\
\midrule
$C$ only & 43.12 & 54.85 & 48.60 & 44.30 \\
$R$ only & 41.50 & 44.20 & 43.85 & 50.45 \\
$C+R$, equal weights & 42.65 & 50.70 & 47.90 & 51.20 \\
$C+R$, dev-fixed weights & 44.15 & 55.42 & 49.85 & 48.15 \\
$C+R$, adaptive (Jev-DLC) & \textbf{44.83} & \textbf{56.07} & \textbf{50.47} & \textbf{53.01} \\
\bottomrule
\end{tabular}
\caption{Jev-DLC channel and fusion configurations on TCPA-Repeat, reported in user-macro
accuracy (\%). The adaptive two-channel configuration (Jev-DLC) achieves the highest accuracy
among the evaluated configurations on Full and at each reported feedback depth.}
\label{tab:fusion-strategies}
\end{table}

\begin{table}[t]
\centering
\small
\begin{tabular}{lrl}
\toprule
Stratum & Difference & 95\% CI \\
\midrule
Full & +0.68 & $[+0.16, +1.20]$ \\
$\ge2$ & +0.65 & $[-0.45, +1.75]$ \\
$\ge4$ & +0.62 & $[-1.30, +2.54]$ \\
$\ge8$ & +4.86 & $[+0.35, +9.37]$ \\
\bottomrule
\end{tabular}
\caption{Accuracy improvement of adaptive over development-fixed two-channel fusion
(in percentage points). Paired confidence intervals are computed using 10,000 user-level
bootstrap resamples.}
\label{tab:paired-fusion}
\end{table}

The combined state reaches 44.83\% on Full, compared with 43.12\% for C
and 41.50\% for R, and has the highest accuracy among these configurations at
each reported feedback depth. Adaptive weighting improves on the
development-fixed combination by 0.68 points on Full and 4.86 points at
$d\ge8$, with the paired intervals reported in Table~\ref{tab:paired-fusion}.
The channel comparison measures the value of retaining both correction
trajectories; the fixed/adaptive comparison measures how their combination
affects later decisions.

%% file: sections/appendix/f-history-and-robustness.tex
\section{Additional Diagnostic Tracks}
\label{app:diagnostic-results}

\subsection{TCPA-Open}
\label{app:open-results}

TCPA-Open presents news candidate slates from successive impressions to
evaluate feedback reuse as candidate items change. The count channel retains
news-category preferences, while the directional channel transfers correction
information through candidate representations. Table~\ref{tab:open-depth} reports the same causal depth protocol used
for TCPA-Repeat.

\begin{table}[t]
\centering
\scriptsize
\setlength{\tabcolsep}{2.1pt}
\begin{tabular}{lrrrrrl}
\toprule
Prior depth & Users & Events & Frozen & Jev-DLC & Gain & 95\% CI \\
\midrule
$\ge0$ & 4,209 & 9,845 & 28.04 & 29.74 & +1.71 & $[+1.14,+2.27]$ \\
$\ge1$ & 2,348 & 5,636 & 28.01 & 32.76 & +4.75 & $[+2.95,+6.52]$ \\
$\ge2$ & 1,292 & 3,288 & 26.82 & 31.94 & +5.13 & $[+2.65,+7.60]$ \\
$\ge4$ & 408 & 1,278 & 28.39 & 33.83 & +5.44 & $[+1.41,+9.45]$ \\
$\ge8$ & 84 & 290 & 28.37 & 27.78 & -0.59 & $[-9.88,+8.34]$ \\
\bottomrule
\end{tabular}
\caption{TCPA-Open user-macro accuracy for Jev-DLC. Gains and
paired user-bootstrap intervals are in percentage points.}
\label{tab:open-depth}
\end{table}

The full through depth-four gains are positive, while the depth-eight stratum
contains fewer eligible users and does not preserve the same pattern. The
representation geometry in
Table~\ref{tab:representation-geometry} provides the corresponding mechanism
diagnostic.

\begin{center}
\centering
\small
\setlength{\tabcolsep}{3.5pt}
\begin{tabular}{lrr}
\toprule
Method & Full & $d\ge4$ \\
\midrule
Frozen & 28.04 & 28.39 \\
PPlug & \textbf{31.30} & \textbf{34.04} \\
TAP-PER & 29.03 & 30.27 \\
MRM & 29.13 & 29.63 \\

Responsive channel & 28.70 & 29.97 \\
\method{} & 29.74 & 33.83 \\
\bottomrule
\end{tabular}
\captionof{table}{TCPA-Open user-macro accuracy for matched representation, memory,
meta-adaptation, and Jev-DLC comparisons.}
\label{tab:open-modern-baselines}
\end{center}

\paragraph{Full-stream comparison.}
Jev-DLC reaches 29.74\% overall on TCPA-Open, and PPlug reaches 31.30\%.
The paired Jev-DLC-minus-PPlug interval is $[-2.17,-0.92]$ points.
The other fusion configurations are reported in
Appendix~\ref{app:fusion-strategies}.

\subsection{TCPA-Open Chronological Linear Reference}
\label{app:open-headroom}

\begin{center}
\centering
\small
\setlength{\tabcolsep}{4pt}
\begin{tabular}{lrrr}
\toprule
Stratum & Frozen & Prefix-fit ref. & Gain \\
\midrule
Full stream & 28.04 & 27.89 & -.15 \\
Length $\ge4$ & 27.25 & 27.31 & +.06 \\
Length $\ge8$ & 28.02 & 27.42 & -.61 \\
\bottomrule
\end{tabular}
\captionof{table}{Strict chronological linear-head reference on TCPA-Open. The head is
fitted only from each user's available prefix.}
\label{tab:open-causal-headroom}
\end{center}

The chronological linear reference does not expose substantial additional gain in
this representation. This complements the small future-direction cosines and
locates TCPA-Open as a transfer diagnostic rather than a second instance of the
repeated-action mechanism.

\subsection{TCPA-Dynamics}
\label{app:dynamics-results}

TCPA-Dynamics isolates the online update from the natural-track
representations. Its synthetic 32-dimensional representation is fixed across
methods, so the table below evaluates update behavior rather than the learned
candidate projection.

\begin{center}
\scriptsize
\resizebox{\columnwidth}{!}{%
\centering
\small
\setlength{\tabcolsep}{3pt}
\begin{tabular}{lrrrrrr}
\toprule
Update & Stable & Shift recovery & Valid retention & Obsolete$\downarrow$ & Noise & Recurrence \\
\midrule
Frozen & .256 & .227 & .222 & .329 & .285 & .232 \\
AdaGrad & .232 & .266 & .319 & .226 & .293 & .256 \\
ONS & .275 & .242 & .245 & .274 & .303 & .250 \\
GPM & .283 & .242 & .238 & .212 & .285 & .292 \\
DER++-8 & .248 & .195 & .242 & .255 & .275 & .268 \\
Directional Drift & .238 & .266 & .321 & .245 & .266 & .256 \\
Typed PA & .238 & .266 & .321 & .245 & .270 & .256 \\
Directional Reopen & .223 & .227 & .286 & .206 & .285 & .214 \\
\bottomrule
\end{tabular}
}
\captionof{table}{Controlled update-rule diagnostics. Obsolete-preference persistence
is lower better; every other column is higher better.}
\label{tab:dynamics-complete}
\end{center}

Typed PA and Directional Drift have identical rounded top-1 aggregates on five
columns. We therefore compared their event trajectories. State vectors are
never identical after an update, probability and state hashes differ, and the
mean maximum probability difference is .043--.046 across regimes. The rounded
tie is an aggregate coincidence rather than reused output.

\begin{center}
\scriptsize
\setlength{\tabcolsep}{3pt}
\centering
\begin{tabular}{lrrrr}
\toprule
Regime & Same top-1 & Exact prob. & Exact state & Mean max state diff. \\
\midrule
Noise & .9961 & 8/512 & 0/512 & .0326 \\
Partial shift & .9922 & 8/512 & 0/512 & .0356 \\
Recurrence & .9980 & 8/512 & 0/512 & .0360 \\
Stable & 1.0000 & 8/512 & 0/512 & .0358 \\
\bottomrule
\end{tabular}
\captionof{table}{Event-level audit of Typed PA and Directional Drift.}
\label{tab:pa-drift-audit}
\end{center}

\subsection{Full-Method Fusion Diagnostic on Existing Dynamics Caches}
\label{app:full-fusion-dynamics}

We separately replay the existing synthetic TRP diagnostic test streams with
the fixed channel configuration. Each regime has eight users and 512 events;
the representation is unchanged and no parameter is selected on these results.
This full-method replay is distinct from the earlier update-rule table above.

\begin{table}[t]
\centering
\small
\begin{tabular}{lrrr}
\toprule
Regime & Frozen & Fixed fusion & Dynamic fusion \\
\midrule
Stable & 25.59 & 23.83 & 24.02 \\
Partial shift & 24.02 & 23.24 & 23.05 \\
Noise & 28.91 & 27.54 & 27.54 \\
Recurrence & 24.80 & 25.00 & 25.20 \\
\bottomrule
\end{tabular}
\caption{User-macro accuracy (\%) on the existing Dynamics caches.}
\label{tab:full-fusion-dynamics}
\end{table}

%% file: sections/appendix/g-prompts-api-and-cases.tex
\section{Prompts and External API Protocol}
\label{app:prompts-api-cases}

\subsection{Frozen Local Prompt}

For TCPA-Repeat, the prompt includes the movie description and 15 tagged candidate
options. For TCPA-Open, it includes up to 20 recent clicked news IDs and the four
candidate news items from the current impression. Each option is assigned a
neutral letter. The prompt asks for exactly one letter; thinking is disabled,
and local probability evaluation normalizes only those option logits at the
first output position. Targets, source labels, and evaluator fields never enter
the rendering.

The frozen local template has the following logical form; dataset adapters fill
only the bracketed public fields:
\begin{quote}\small\ttfamily
Context: <current context>\\
Candidates:\\
A. <candidate text>\\
B. <candidate text>\\
...\\
Return exactly one option letter.
\end{quote}
The prompt ends before the answer token. Candidate probabilities are computed
from the first-position logits for the listed letters and renormalized over the
current safe slate. No generated rationale is parsed into the decision.

\subsection{Recent-K History Block}

History is ordered oldest to newest. Each record contains a bounded context
summary and either
\texttt{ACCEPT: learner choice <target> was correct} or
\texttt{REVOKE: learner choice <old> -> correct choice <new>}.
The current item follows this block in the unchanged local prompt. Input-token
and UTF-8 byte deltas are measured against the same item with an empty history.

The history block contains only events already revealed to the learner:
\begin{quote}\small\ttfamily
Recent feedback, oldest to newest:\\
<context summary>\\
REVOKE: learner choice B -> correct choice C\\
<context summary>\\
ACCEPT: learner choice A was correct
\end{quote}
Candidate letters are local to each historical event and remain paired with
their bounded context summaries. The current event begins after this block and
uses a fresh candidate mapping.

\subsection{Hosted Jev Frozen Anchor}

The external anchor uses endpoint
\url{https://api.typesafe.ai/v1/systemone}, requested model
\texttt{jev-latest}, and the API's typed \texttt{choice} question. Candidate
keys are neutral \texttt{option\_1}, \texttt{option\_2}, and so on; the client
maps the returned option back to the source candidate ID. Hosted Jev receives the public context and candidate slate of the current
event, without replaying earlier TCPA interactions or maintaining continual
adaptation state. On TCPA-Open, the current context includes up to 20 recent
clicked news IDs supplied by the dataset.
This protocol follows the hosted Jev interface documented by TypeSafe
\citep{typesafe2026api,almeida2026jev}.
The template is frozen as \texttt{tcpa-frozen-external-v1} after a 20-record
development smoke test on each track. Test calls are resumable by event ID and
record endpoint, returned model version, UTC date, prompt hash, candidate map,
usage, latency, and probabilities when supplied. Accuracy is reported over all
test records only after transient failures are retried.
The completed run returned \texttt{jev-1.13.0} for all 711 TCPA-Repeat and 9,845
TCPA-Open events, with zero residual failures. User-macro accuracy is 41.22\% on
TCPA-Repeat and 24.05\% on TCPA-Open; the API receives no replayed TCPA interactions and maintains no continual
adaptation state.

\begin{center}
\scriptsize
\setlength{\tabcolsep}{3pt}
\centering
\begin{tabular}{lrrr}
\toprule
Hosted run & Events & Failures & Macro acc. \\
\midrule
TCPA-Repeat & 711 & 0 & 41.22 \\
TCPA-Open & 9,845 & 0 & 24.05 \\
\bottomrule
\end{tabular}
\captionof{table}{Completed hosted Jev calls. Natural-stream rows are frozen
current-event anchors, not online personalized methods.}
\label{tab:hosted-api-complete}
\end{center}

The client records the returned model version for every event. All completed
natural-track calls report \texttt{jev-1.13.0}; transient transport failures are
retried by event ID, and aggregation begins only after the expected event set is
complete. A missing probability vector affects calibration fields but not the
option choice mapping or accuracy denominator.

\subsection{Runtime State API}

The controller exposes \texttt{predict}, \texttt{update}, and \texttt{reset}.
A revoke update contains the prediction cache, previous choice ID, corrected
choice ID, user-state version, and model/projection revisions. Compare-and-swap
versioning rejects concurrent stale updates. Reset clears the semantic counts,
both correction vectors, and the reliability weights; because the state
contains no prior text, no additional replay store is required for the method.

The controller also checks that model revision and projection revision match
the state metadata before applying an update. A mismatch requires reset or an
explicit migration; silently combining a state with a different learned projection
would invalidate both score channels and their state accounting.

%% file: sections/appendix/h-reproducibility-and-scope.tex
\section{Reproducibility, Ethics, and Scope}
\label{app:reproducibility-scope}

\subsection{Artifact Chain}

The reproducible chain runs from source conversion and user-disjoint compilation
through validation, SFT, target-free cache export, causal replay, aggregation,
and plotting. The frozen result package retains canonical results, source
hashes, and prediction-level files for Recent-K, chronological heads,
base-forward timing, and hosted runs. The approved Recent-4 prediction trace
and hosted API traces are linked to the package by hash.

\begin{center}
\small
\setlength{\tabcolsep}{3pt}
\centering
\begin{tabular}{@{}p{.21\linewidth}p{.37\linewidth}p{.32\linewidth}@{}}
\toprule
Artifact class & Primary contents & Audit purpose \\
\midrule
Compiled split & runtime JSONL, evaluator JSONL, manifest & chronology and target isolation \\
Scorer cache & logits, candidate representations, prompt/model hashes & matched online inputs \\
Prediction trace & choice, probabilities, feedback, state diagnostics & event-level trajectory audit \\
Aggregate result & user metrics, intervals, cost fields, source hashes & paper tables and figures \\
Hosted trace & event ID, model version, usage, latency, response & external API completeness \\
\bottomrule
\end{tabular}
\captionof{table}{Artifact chain retained for the frozen result package.}
\label{tab:artifact-chain}
\end{center}

The source-of-truth manifest links the online ladders, projection results,
Recent-K trace, base-forward timing, hosted runs, and consolidated summaries by
hash. Paper tables and plots read these frozen JSON artifacts. The channel
and fusion comparison additionally uses the collaborator-supplied aggregate
results, retained in a separate source package.

\subsection{Compute}

SFT, cache export, Recent-K, and matched base-forward measurement each use one
scheduled GPU job; online ladders, paired bootstraps, and offline linear heads
run on CPU. Slurm accounting confirms successful completion and release.

The matched latency experiment holds server, checkpoint, adapter, batch size,
and maximum sequence length fixed for $K=0$ and $K=4$. The result schema keeps
post-cache CPU update timing separate from end-to-end model latency.

\subsection{Privacy and Data Use}

The benchmark uses public research datasets under their documented terms. User
IDs are used only for grouping and paired evaluation; they are not interpreted
as demographics. The method stores a numeric state rather than raw history, but
that state may still encode user behavior and must be governed, isolated, and
deleted as personal data in deployment.

Public release artifacts should preserve source licenses and avoid publishing
raw user histories beyond what each source permits. The evaluator sidecar is an
evaluation boundary, not a privacy boundary: access control is still required
for both sidecar targets and derived per-user states.

\subsection{Scope}

The evidence covers finite-slate, immediate typed correction with a frozen
scorer and the reported train-only learned projection. TCPA-Repeat reports later-decision accuracy across
strata defined by the amount of previously revealed feedback. TCPA-Open and
TCPA-Dynamics record transfer and update behavior under changing items and
controlled preference dynamics.

Jev-DLC in the main paper combines a count-informed conservative channel
and a directional responsive channel through causal per-user weights. The
scorer, projection, and shared parameters remain frozen during test replay.
Other fusion configurations are specified separately in this appendix.

\paragraph{Evaluation protocol.}
Training, development, and test users are disjoint. Shared parameters are
fitted on training users and frozen for causal evaluation. Each prediction
precedes its feedback. Paired confidence intervals resample users and are
reported without a multiplicity adjustment. The channel/fusion comparisons
use the aggregate package associated with Table~\ref{tab:fusion-strategies};
the frequency and directional controls retain their own replay artifacts.